\documentclass{article}
\usepackage{iclr2026_conference,times}

\usepackage{hyperref}
\usepackage{url}
\usepackage{graphicx}
\usepackage{booktabs}
\usepackage{amsmath}
\usepackage{amssymb}
\usepackage{float}
\usepackage{subcaption}
\usepackage[most]{tcolorbox}
\usepackage{caption}
\newtcolorbox{userstory}[1]{
  colbacktitle=black,
  coltitle=white,
  fonttitle=\bfseries,
  colback=white,
  colframe=black,
  title=#1,
  boxrule=1.2pt,
  arc=4pt,
  outer arc=4pt,
  top=4pt,
  bottom=4pt,
  left=4pt,
  right=4pt
}

\newtcolorbox{primitivebox}[2][]{
  colback=white,
  colframe=gray!40,
  colbacktitle=gray!15,
  coltitle=black,
  fonttitle=\bfseries\small,
  title=#2,
  boxrule=0.8pt,
  arc=4pt,
  outer arc=4pt,
  top=2pt,
  bottom=2pt,
  left=4pt,
  right=4pt,
  toptitle=1pt,
  bottomtitle=1pt
}

\title{GeoAI Agency Primitives}

\author{
Akram Zaytar$^{1,\ast}$ \quad Rohan Sawahn$^{2}$ \quad Caleb Robinson$^{1}$ \quad Gilles Q. Hacheme$^{1}$ \\
Girmaw A. Tadesse$^{1}$ \quad Inbal Becker-Reshef$^{1}$ \quad Rahul Dodhia$^{1}$ \quad Juan Lavista Ferres$^{1}$ \\
$^{1}$Microsoft AI for Good Lab \quad $^{2}$NASA Harvest \\
$^{\ast}$Corresponding author: \texttt{akramzaytar@microsoft.com} \\
}

\iclrfinalcopy
\begin{document}

\maketitle

\begin{abstract}
We present ongoing research on agency primitives for GeoAI assistants---core capabilities that connect Foundation models to the artifact-centric, human-in-the-loop workflows where GIS practitioners actually work. Despite advances in satellite image captioning, visual question answering, and promptable segmentation, these capabilities have not translated into productivity gains for practitioners who spend most of their time producing vector layers, raster maps, and cartographic products. The gap is not model capability alone but the absence of an agency layer that supports iterative collaboration. We propose a vocabulary of $9$ primitives for such a layer---including navigation, perception, geo-referenced memory, and dual modeling---along with a benchmark that measures human productivity. Our goal is a vocabulary that makes agentic assistance in GIS implementable, testable, and comparable.
\end{abstract}

\section{Introduction}

The most used AI assistants today augment human intelligence rather than replace it. GitHub Copilot helps programmers write code faster \citep{peng2023copilot}, but the programmer still architects, tests, and quality-controls the solution. ChatGPT helps writers draft text, but the writer still decides what to say. These tools succeed because they operate in human-intelligible formats --- code and text ---where people can inspect, edit, and iterate on suggestions. Geographic Information Systems (GIS) present a more challenging case. Practitioners spend most of their time producing artifacts: vector layers, raster maps, and infographic and cartographic products. The tasks are repetitive yet the work is careful, controlled, and iterative---digitizing boundaries, labeling land cover, correcting misalignments, validating outputs---and the output space is spatial, temporal, and visual.

This multimodal artifact-centric nature of GIS work creates challenges for GeoAI automation. GIS artifacts are not text-native. We need extensive scaffolding and interfaces to make large language models (LLMs) and vision–language models (VLMs) work for GIS artifacts. On top of that, the earth cannot fit into an LLM's context window. Yet, most GIS problems are local: a farmer cares about their agricultural land, not global crop statistics; a city planner needs to map a neighborhood's flood risk, not continental averages. This presents a great opportunity. Instead of trying to build a single model that works for everything, everywhere, we should build the scaffolding that lets people use powerful models to create solutions tailored to the places and problems they care about.

\begin{figure}[htbp]
  \centering
  \includegraphics[width=0.7\linewidth]{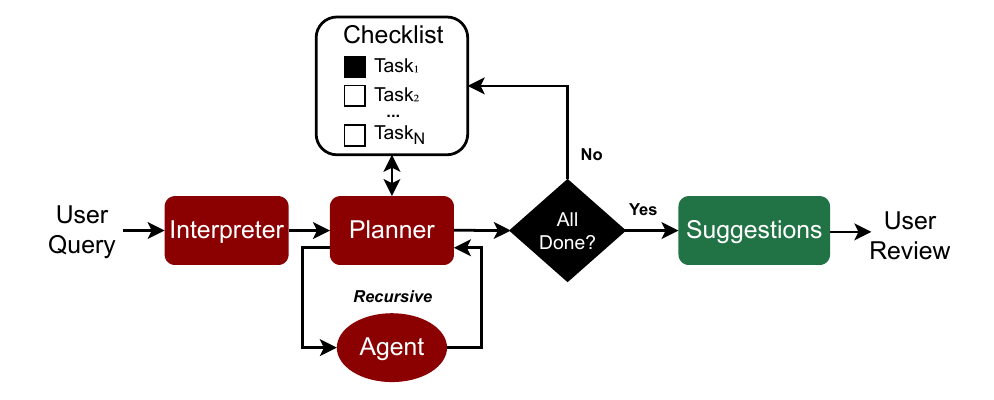}
  \caption{\textbf{General agent workflow.} User queries are parsed, decomposed into tasks, and executed recursively. Completed tasks yield suggestions for user review before committing.}
  \label{fig:overview}
\end{figure}

A growing body of work connects language models to GIS tools through tool-calling, code-generation, and hybrid approaches (see Table~\ref{tab:capability-matrix} in Appendix). Despite this progress, existing systems function primarily as natural language interfaces for GIS \textbf{knowledge extraction}---they do not support the human-in-the-loop, iterative, and GIS-native workflows that artifact production requires. No existing system implements a GeoAI agency framework for \textbf{Dataset Development} with vision-native suggest-review-commit interaction (Figure~\ref{fig:overview}), context navigation, geo-memory, compute budgets, or model-tiers for label propagation.

We propose a framework of agency primitives that define how an agent navigates imagery, perceives and describes scenes, remembers what it learned about a place, proposes edits for user review, and scales sparse supervision to full-coverage maps. Instead of measuring accuracy alone, we propose a benchmarking framework capturing human productivity through time-to-threshold, progress curves, rework rate, and suggestion bias. We illustrate how these primitives compose through user stories across crop mapping, disaster assessment, and image summarization. We present this as a conceptual framework for community discussion; implementation and validation remain future work.

\section{Agency Primitives}\label{sec:primitives}

An agency primitive is a capability that lets an assistant take grounded actions in a GIS workspace while keeping the user in control. Primitives are not models---they are interfaces or tools that enable human-AI collaboration on GIS tasks. We describe nine primitives: Navigation, Perception, Geo-Memory, Embeddings, Graphs, Budgets, Propagation, Attribution, and Dual Modeling.

\begin{figure}[t!]
  \centering
  \begin{subfigure}[t]{0.48\linewidth}
    \begin{primitivebox}[sensing]{Navigation}
      \centering
      \includegraphics[width=\linewidth]{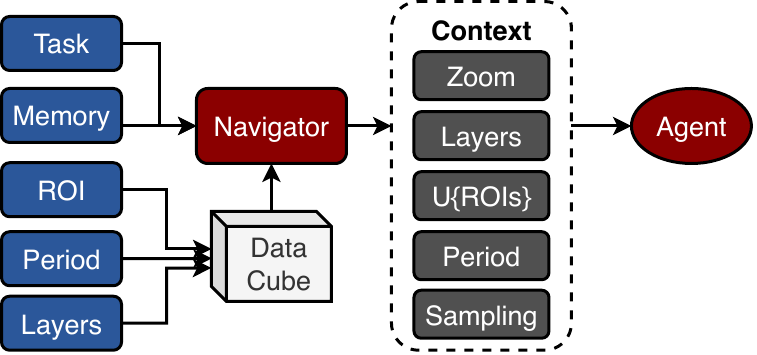}
    \end{primitivebox}
  \end{subfigure}
  \hfill
  \begin{subfigure}[t]{0.48\linewidth}
    \begin{primitivebox}[sensing]{Perception}
      \centering
      \includegraphics[width=\linewidth]{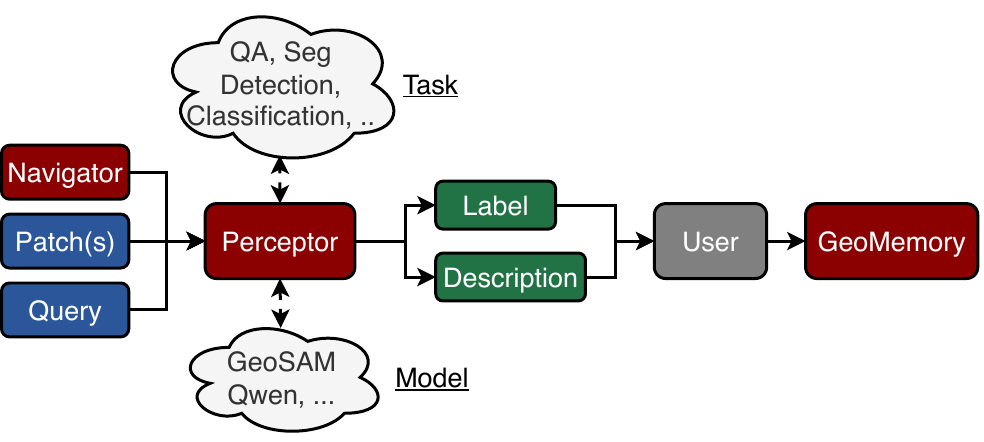}
    \end{primitivebox}
  \end{subfigure}
  \\[4pt]
  \begin{subfigure}[t]{0.48\linewidth}
    \begin{primitivebox}[sensing]{Georeferenced Memory}
      \centering
      \includegraphics[width=\linewidth]{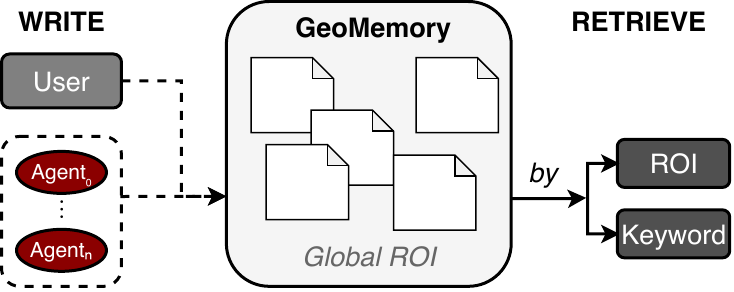}
    \end{primitivebox}
  \end{subfigure}
  \hfill
  \begin{subfigure}[t]{0.48\linewidth}
    \begin{primitivebox}[sensing]{Earth Embeddings}
      \centering
      \includegraphics[width=\linewidth]{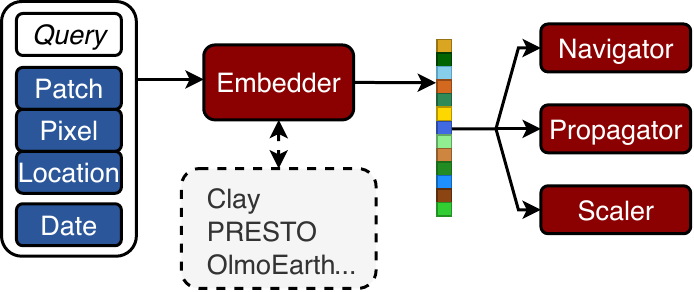}
    \end{primitivebox}
  \end{subfigure}
  \caption{\textbf{Core sensing primitives.} (Top-left) Navigation constructs context bundles specifying sub-ROIs, zoom, and sampling strategy. (Top-right) Perception routes patches to task-appropriate models, returning labels and notes. (Bottom-left) GeoMemory stores spatial notes for retrieval and curation. (Bottom-right) Embeddings map inputs to vectors for similarity search and modeling.}
  \label{fig:sensing_primitives}
\end{figure}

\paragraph{Grounded Navigation:} Geospatial data is too large to feed to a VLM at once. A single Sentinel-2 tile covers roughly 100 km $\times$ 100 km. Navigation is how the agent decides what to look at. Given a user query and workspace state, the navigator produces a spatio-temporal-layer context bundle (Figure~\ref{fig:sensing_primitives}, top-left): which sub-regions to examine, at what times, at what zoom level, with which layer views (band combinations), and by which sampling strategy. The choice of a sampling strategy depends on the task: exploration favors diversity; quality control favors uncertainty; systematic mapping favors coverage; change detection favors temporal contrast.

\paragraph{Perception:} Once the Navigator sets appropriate context, it needs to ``see'' it. Perception is an interface that routes queries to task-appropriate models: object detection (YOLTv5, MMRotate), segmentation (SAM, DeepForest), visual question answering (GeoChat, RemoteCLIP), or change detection (ChangeFormer) \citep{kirillov2023sam, weinstein2019individual, geochat2024, remoteclip2023, bandara2022transformer}. For each patch, the Perceptor receives $\ge 1$ patches, geographic metadata, and a task query. It returns a label and a ``note'' describing what it observed. When the perceptor cannot answer a query because the resolution is too low, the view is occluded, or the scene is ambiguous then it can use the ``note'' space to say so and explain why. This prevents silent failures and enables targeted follow-up. Perception outputs are suggestions by default (Figure~\ref{fig:sensing_primitives}, top-right).

\paragraph{GeoMemory:} Memory makes continual learning (i.e., accumulated understanding of a place) and iterative work possible. Concretely, memory is a blank canvas representing the global ROI that accumulates perceptor snapshots as polygons; each entry stores (\texttt{geometry, timestamp, query, output reference}, and \texttt{notes)} and can later be queried by spatial intersection, time ranges, or keywords. Memory supports three operations (Figure~\ref{fig:sensing_primitives}, bottom-left). ``WRITE'' adds or updates entries with a location, timestamp, and content. ``RETRIEVE'' fetches relevant entries by spatial query, temporal window, or keyword search. ``CURATE'' lets users delete, correct, or confirm entries in ``suggestion'' mode. Efficient retrieval requires spatial (R-tree) and temporal indexing.

\begin{figure}[t!]
  \centering
  \begin{subfigure}[t]{0.48\linewidth}
    \begin{primitivebox}[execution]{Compute Graphs}
      \centering
      \includegraphics[width=\linewidth]{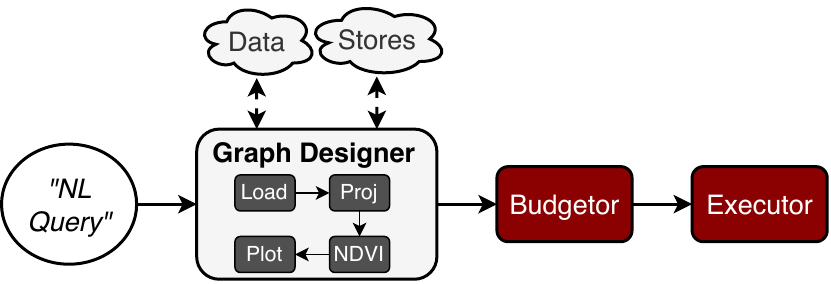}
    \end{primitivebox}
  \end{subfigure}
  \hfill
  \begin{subfigure}[t]{0.48\linewidth}
    \begin{primitivebox}[execution]{Compute Budgets}
      \centering
      \includegraphics[width=\linewidth]{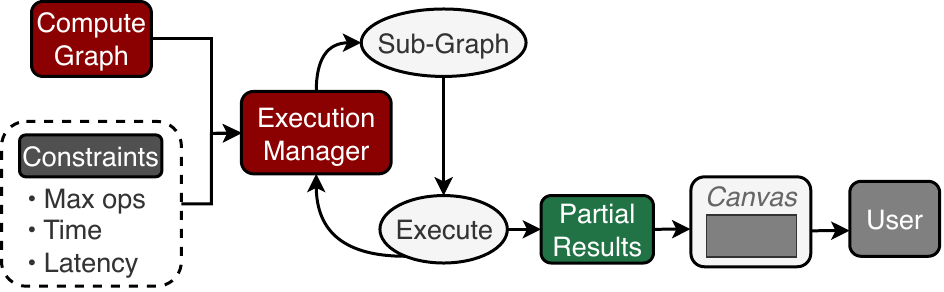}
    \end{primitivebox}
  \end{subfigure}
  \\[4pt]
  \begin{subfigure}[t]{0.48\linewidth}
    \begin{primitivebox}[execution]{Attribution}
      \centering
      \includegraphics[width=\linewidth]{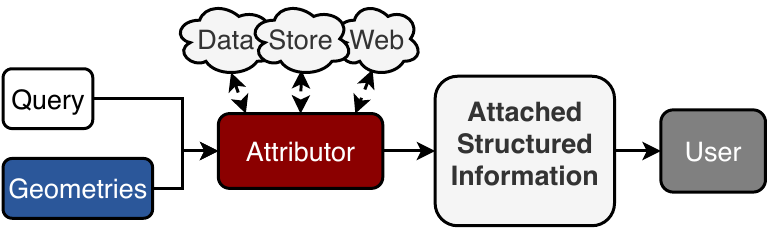}
    \end{primitivebox}
  \end{subfigure}
  \hfill
  \begin{subfigure}[t]{0.48\linewidth}
    \begin{primitivebox}[execution]{Dual Modeling}
      \centering
      \includegraphics[width=\linewidth]{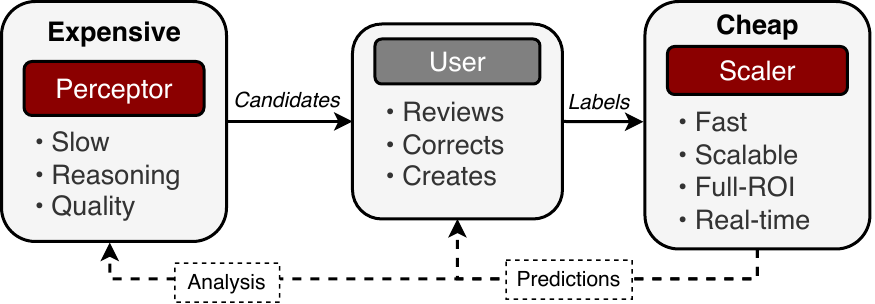}
    \end{primitivebox}
  \end{subfigure}
  \caption{\textbf{Execution and enrichment primitives.} (Top-left) Compute Graphs translate queries into directed operation graphs. (Top-right) Budgets enforce constraints enabling partial results and early stopping. (Bottom-left) Attribution enriches geometries with external data. (Bottom-right) Dual Modeling iterates between expensive VLM judgments and cheap scalable inference.}
  \label{fig:execution_primitives}
\end{figure}

\paragraph{Earth Embeddings:} Embeddings provide a semantic layer for similarity search, sampling, and modeling. An embedding maps a patch, pixel, or location to a fixed-length vector capturing its semantics. The embedding interface routes to foundation models appropriate for the input modality and task (Figure~\ref{fig:sensing_primitives}, bottom-right): PRESTO for agricultural time series~\citep{presto2024}, SatCLIP for location-aware priors~\citep{satclip2023}, or Prithvi for Landsat/HLS data~\citep{prithvi2023}. Embeddings serve three roles. First, they enable diversity sampling for navigation---selecting patches spread out in embedding space ensures variety. Second, they power guided propagation---finding examples similar to user-provided seeds. Third, they act as features for lightweight models that scale sparse labels to full-ROI predictions.

\paragraph{Compute Graphs \& Budgets:} A \textbf{Compute Graph} represents computation as a directed graph of operations over workspace layers (Figure~\ref{fig:execution_primitives}, top-left). Nodes represent operations while edges indicate data flow. The graph is inspectable before execution and produces artifacts with provenance. Graph construction can be initiated by natural language (``compute NDVI for my polygons and show the time series'') or by an agent. Furthermore, \textbf{Budgets} are constraints on graph execution that ensure interactivity. They limit how much computation a node can perform (e.g., maximum graph size, operations, or number of VLM calls). If a node exceeds its budget, it is broken into sequential steps with intermediate outputs. Budgets make large-scale processing manageable and allow users to review partial results and decide whether to continue, adjust, or stop early.

\paragraph{Propagation:} Propagation expands user labels by suggesting new candidates. It operationalizes ``find more like this'' in a GIS-native way. Given selected seed labels with attributes and an embedding space, propagation returns ranked candidates: locations similar to the seeds. Candidates appear as suggestions with similarity scores. Users can batch-accept/reject or review. Propagation is local and interactive---it accelerates label collection rather than replacing it. Our goal is rapid discovery of positives, not automated classification (see Figure~\ref{fig:execution_primitives}, bottom-right, and~\ref{fig:story_crop_mapping} for propagation in context).

\paragraph{Attribution:} The Attributor adds information to a selected geometry (Figure~\ref{fig:execution_primitives}, bottom-left) to help the user better understand the geometry or make labeling decisions. Attributes can be textual (e.g., web search findings, mined statistics), imagery (e.g., temporal overviews), categories (e.g., OSM tags, census demographics), or plots (e.g., weather or vegetation signals from ERA5). Computed attributes are derived on-the-fly: zonal statistics from underlying rasters (NDVI, elevation range), shape metrics (area, compactness), or spectral indices extracted for the polygon's footprint.

\paragraph{Dual Modeling:} Dual modeling enables real-time work by iterating between expensive label mining and cheap scalable inference (Figure~\ref{fig:execution_primitives}, bottom-right). Human and perceptor work is slow and expensive but produces good quality outputs. It excels at tasks requiring reasoning: interpreting ambiguous scenes, explaining failures, proposing edits. On the other hand, lightweight models---random forests using embeddings---are cheap to run at scale but lack contextual understanding and fail on out-of-distribution inputs. Dual modeling enables iterating between both. The perceptor provides seed labels, the user diagnoses errors, and handles edge cases. The lightweight model scales. The user closes the loop by reviewing outputs, correcting mistakes, and iterating. This division mirrors successful human-in-the-loop labeling systems: an expert decides what to label and how to fix errors; automation handles the volume.

\section{Benchmark Proposal}\label{sec:benchmark}

We propose a benchmarking framework where $N$ users complete $M$ GIS work sessions, each focused on solving a specific task in a bounded spatio-temporal domain. Sessions sample from three dimensions: a task space $\mathcal{T}$ from existing RS benchmarks (e.g., GEO-Bench, SustainBench), a region $\mathcal{S}$ within task coverage, and a period $\mathcal{W}$ when applicable. The combinatorial space $(t, s, w) \sim \mathcal{T} \times \mathcal{S} \times \mathcal{W}$ ensures near-unique sessions that prevent memorization while reusing existing labels. Each session follows a lifecycle: (1) sample $(t, s, w)$ with all reference data withheld; (2) work interactively while a background evaluator computes task-dependent quality $Q(t)$ (F1 for classification, IoU for segmentation, etc.) against held-out labels at fixed intervals; (3) stop when the user declares ``done'' or a time budget $T_{\max}$ is reached; (4) compute final metrics (Table~\ref{tab:benchmark-metrics}) and store full interaction logs. To quantify the impact of each primitive, we compare four capability levels: \emph{Baseline} (manual labeling tools only), \emph{+Propagation} (adds Guided Propagation and Embeddings), \emph{+Scaling} (adds Dual Modeling and Compute Graphs), and \emph{+Agent} (full stack).

\begin{table}[t]
\centering
\small
\begin{tabular}{@{}ll@{}}
\toprule
\textbf{Metric} & \textbf{Definition} \\
\midrule
Time-to-threshold & $T_\tau = \min\{t : Q(t) \geq \tau\}$, time to reach quality $\tau$ \\
Progress AUC & $\frac{1}{T}\int_0^T Q(t)\,dt$, normalized area under quality curve \\
Rework rate & $R = n_{\text{overwrite}} / n_{\text{edits}}$, fraction of edits that revise prior work \\
Suggestion bias & $\|P_{\text{err}}^{\text{accept}} - P_{\text{err}}^{\text{gt}}\|$, error distribution shift from suggestions \\
\bottomrule
\end{tabular}
\caption{The benchmark tracks four primary metrics. In addition, accept/reject rates, compute cost, and GIS validity (geometry, CRS, schema) are logged for further evaluation.}
\label{tab:benchmark-metrics}
\end{table}

\section{Discussion}\label{sec:future_work}

The most realistic adoption path for GIS agency is incremental: start with visible, immediate wins like faster labeling, better evidence attachments, and safer edits in familiar GIS software rather than promising full autonomy. The primitives and workflows described in this paper are a starting point towards the implementation of such a GeoAI agency framework. Our next steps are to open-source the playground and measure which capabilities reduce time-to-acceptable-artifacts, which GIS tasks benefit most, how users trade off control versus automation, and how agents can learn from failures.

\bibliography{references}

@misc{geoagent2024,
  title         = {{An LLM Agent for Automatic Geospatial Data Analysis}},
  author        = {Chen, Yuxing and Wang, Weijie and Lobry, Sylvain and Kurtz, Camille},
  year          = {2024},
  eprint        = {2410.18792},
  archivePrefix = {arXiv},
  primaryClass  = {cs.AI},
  url           = {https://arxiv.org/abs/2410.18792}
}

@article{giscopilot2025,
  title   = {{GIS Copilot: Towards an Autonomous GIS Agent for Spatial Analysis}},
  author  = {Akinboyewa, Temitope and Li, Zhenlong and Ning, Huan},
  journal = {International Journal of Digital Earth},
  volume  = {18},
  number  = {1},
  year    = {2025},
  doi     = {10.1080/17538947.2025.2497489}
}

@inproceedings{geollmengine2024,
  title     = {{GeoLLM-Engine: A Realistic Environment for Building Geospatial Copilots}},
  author    = {Singh, Simranjit and Fore, Michael and Stamoulis, Dimitrios},
  booktitle = {Proceedings of the IEEE/CVF Conference on Computer Vision and Pattern Recognition Workshops (CVPRW), EarthVision Workshop},
  year      = {2024}
}

@misc{geollmsquad2025,
  title         = {{Multi-Agent Geospatial Copilots for Remote Sensing Workflows}},
  author        = {Lee, Chaehong and Paramanayakam, Varatheepan and Karatzas, Andreas and others},
  year          = {2025},
  eprint        = {2501.16254},
  archivePrefix = {arXiv},
  primaryClass  = {cs.AI},
  url           = {https://arxiv.org/abs/2501.16254}
}

@misc{geogpt2023,
  title         = {{GeoGPT: Understanding and Processing Geospatial Tasks through An Autonomous GPT}},
  author        = {Zhang, Yifan and Wei, Cheng and Wu, Shangyou and He, Zhengting and Yu, Wenhao},
  year          = {2023},
  eprint        = {2307.07930},
  archivePrefix = {arXiv},
  primaryClass  = {cs.CL},
  url           = {https://arxiv.org/abs/2307.07930}
}

@article{autonomousgis2023,
  title   = {{Autonomous GIS: the next-generation AI-powered GIS}},
  author  = {Li, Zhenlong and Ning, Huan},
  journal = {International Journal of Digital Earth},
  volume  = {16},
  number  = {2},
  pages   = {4668--4686},
  year    = {2023},
  publisher = {Taylor \& Francis}
}

@article{mapgpt2024,
  title   = {{MapGPT: An Autonomous Framework for Mapping by Integrating Large Language Model and Cartographic Tools}},
  author  = {Zhang, Yifan and He, Zhengting and Li, Jingxuan and Lin, Jianfeng and Guan, Qingfeng and Yu, Wenhao},
  journal = {Cartography and Geographic Information Science},
  year    = {2024},
  doi     = {10.1080/15230406.2024.2404868},
  pages   = {717--743},
  volume  = {51},
  number  = {6}
}

@misc{treegpt2023,
  title         = {{Tree-GPT: Modular Large Language Model Expert System for Forest Remote Sensing Image Understanding and Interactive Analysis}},
  author        = {Du, Siqi and Tang, Shengjun and Wang, Weixi and Li, Xiaoming and Guo, Renzhong},
  year          = {2023},
  eprint        = {2310.04698},
  archivePrefix = {arXiv},
  primaryClass  = {cs.CV},
  url           = {https://arxiv.org/abs/2310.04698}
}

@article{peng2023copilot,
  title={The Impact of {AI} on Developer Productivity: Evidence from {GitHub Copilot}},
  author={Peng, Sida and Kalliamvakou, Eirini and Cihon, Peter and Demirer, Mert},
  journal={arXiv preprint arXiv:2302.06590},
  year={2023}
}

@inproceedings{presto2024,
  title         = {{Lightweight, Pre-trained Transformers for Remote Sensing Timeseries}},
  author        = {Tseng, Gabriel and Cartuyvels, Ruben and Zvonkov, Ivan and Purohit, Mirali and Rolnick, David and Kerner, Hannah},
  booktitle     = {NeurIPS 2023 Workshop on Tackling Climate Change with Machine Learning},
  year          = {2023},
  eprint        = {2304.14065},
  archivePrefix = {arXiv},
  primaryClass  = {cs.CV},
  url           = {https://arxiv.org/abs/2304.14065}
}

@misc{prithvi2023,
  title         = {{Foundation Models for Generalist Geospatial Artificial Intelligence}},
  author        = {Jakubik, Johannes and others},
  year          = {2023},
  eprint        = {2310.18660},
  archivePrefix = {arXiv},
  primaryClass  = {cs.CV},
  url           = {https://arxiv.org/abs/2310.18660}
}

@misc{satclip2023,
  title         = {{SatCLIP: Global, General-Purpose Location Embeddings with Satellite Imagery}},
  author        = {Klemmer, Konstantin and Rolf, Esther and Robinson, Caleb and Mackey, Lester and Ru{\ss}wurm, Marc},
  year          = {2023},
  eprint        = {2311.17179},
  archivePrefix = {arXiv},
  primaryClass  = {cs.CV},
  url           = {https://arxiv.org/abs/2311.17179}
}

@article{remoteclip2023,
  title   = {{RemoteCLIP: A Vision Language Foundation Model for Remote Sensing}},
  author  = {Liu, Fan and Chen, Delong and Guan, Zhangqingyun and Zhou, Xiaocong and Zhu, Jiale and Ye, Qiaolin and Fu, Liyong and Zhou, Jun},
  journal = {IEEE Transactions on Geoscience and Remote Sensing},
  volume  = {62},
  pages   = {1--16},
  year    = {2024},
  publisher = {IEEE}
}

@inproceedings{geochat2024,
  title         = {{GeoChat: Grounded Large Vision--Language Model for Remote Sensing}},
  author        = {Kuckreja, Kartik and Danish, Muhammad Sohail and Naseer, Muzammal and Das, Abhijit and Khan, Salman and Khan, Fahad Shahbaz},
  booktitle     = {Proceedings of the IEEE/CVF Conference on Computer Vision and Pattern Recognition (CVPR)},
  year          = {2024},
  eprint        = {2311.15826},
  archivePrefix = {arXiv},
  primaryClass  = {cs.CV},
  url           = {https://arxiv.org/abs/2311.15826}
}

@misc{kirillov2023sam,
  title         = {{Segment Anything}},
  author        = {Kirillov, Alexander and Mintun, Eric and Ravi, Nikhila and Mao, Hanzi and Rolland, Chloe and Gustafson, Laura and Xiao, Tete and Whitehead, Spencer and Berg, Alexander C. and Lo, Wan-Yen and Doll{\'a}r, Piotr and Girshick, Ross},
  year          = {2023},
  eprint        = {2304.02643},
  archivePrefix = {arXiv},
  primaryClass  = {cs.CV},
  url           = {https://arxiv.org/abs/2304.02643}
}

@inproceedings{bandara2022transformer,
  title={A transformer-based siamese network for change detection},
  author={Bandara, Wele Gedara Chaminda and Patel, Vishal M},
  booktitle={IGARSS 2022-2022 IEEE International Geoscience and Remote Sensing Symposium},
  pages={207--210},
  year={2022},
  organization={IEEE}
}

@article{weinstein2019individual,
  title={Individual tree-crown detection in RGB imagery using semi-supervised deep learning neural networks},
  author={Weinstein, Ben G and Marconi, Sergio and Bohlman, Stephanie and Zare, Alina and White, Ethan},
  journal={Remote Sensing},
  volume={11},
  number={11},
  pages={1309},
  year={2019},
  publisher={MDPI}
}
\bibliographystyle{iclr2026_conference}

\clearpage

\appendix

\section{Capability Comparison}\label{sec:capability-comparison}

\begin{table}[h]
\centering
\footnotesize
\setlength{\tabcolsep}{3pt}
\begin{tabular}{@{}llcccc@{}}
\toprule
\textbf{System} & \textbf{Approach} & \textbf{Suggest-Review} & \textbf{GeoMemory} & \textbf{Budgets} & \textbf{Propag.} \\
\midrule
GeoGPT \citep{geogpt2023} & Tool-call & --- & --- & --- & --- \\
AutoGIS \citep{autonomousgis2023} & Code-gen & --- & --- & --- & --- \\
MapGPT \citep{mapgpt2024} & Tool-call & partial & --- & --- & --- \\
Tree-GPT \citep{treegpt2023} & Hybrid & --- & --- & --- & --- \\
GIS Copilot \citep{giscopilot2025} & Tool+code & partial & --- & --- & --- \\
GeoAgent \citep{geoagent2024} & Code+MCTS & --- & --- & partial & --- \\
GeoLLM-Eng. \citep{geollmengine2024} & Tool-aug. & --- & --- & --- & --- \\
GeoLLM-Sq. \citep{geollmsquad2025} & Multi-agent & --- & --- & partial & --- \\
\bottomrule
\end{tabular}
\caption{Capability comparison of GeoAI agent frameworks. Existing systems span three paradigms: \textbf{Tool-calling} uses LangChain-style agents with predefined function pools; \textbf{Code-generation} has LLMs produce Python scripts directly; \textbf{Hybrid} systems combine multiple approaches. Columns indicate: Suggest-Review (iterative human approval before commit), Georef.\ Memory (spatially-indexed state across sessions), Budgets (constraints on compute/time/cost), Propagation (``find more like this'' for efficient labeling).}
\label{tab:capability-matrix}
\end{table}

\section{User Stories}\label{sec:stories}

We illustrate how our primitives compose through three user stories: image summarization, crop mapping with sparse labels, and flood damage assessment.

\begin{userstory}{Image Summarization}
\centering
\includegraphics[width=\linewidth]{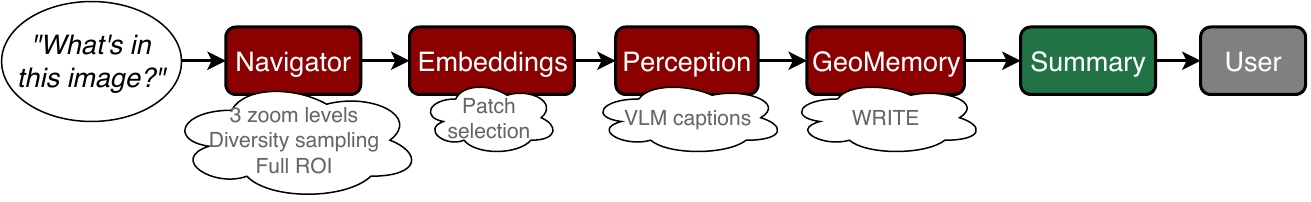}
\label{fig:story_image_summarization}

\vspace{0.5em}
\raggedright
A user loads a satellite image and asks: ``What's in this image?'' The agent interprets this as a multi-scale exploration task. It selects three zoom levels and uses diversity sampling over embeddings to pick representative patches at each scale. A VLM captions each patch, and the results are written to geo-memory. Once complete, the agent synthesizes a summary: ``Agricultural land in the south with small field parcels. A town in the northeast with dense low-rise buildings. Forested hills in the northwest. Mostly residential buildings.'' The user can now ask follow-up questions (``Tell me more about the agricultural area'') or pivot to labeling (``Help me map the maize fields''). The initial summarization seeds geo-memory with observations that accelerate subsequent work.
\end{userstory}

\begin{userstory}{Crop Mapping with Sparse Labels}
\centering
\includegraphics[width=\linewidth]{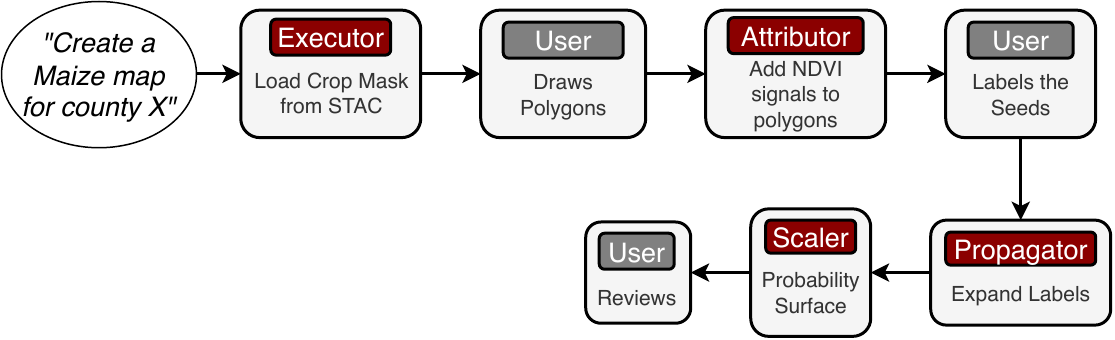}
\label{fig:story_crop_mapping}

\vspace{0.5em}
\raggedright
A user wants a maize probability map for a Kenyan county with minimal manual labeling. They set the ROI to the county boundary and the time period to the growing season. The agent first loads a cropland mask, narrowing the problem from ``maize vs.\ everything'' to ``maize vs.\ other crops.'' The user draws candidate field polygons but cannot confidently label them from imagery alone. The agent builds a compute graph that attaches NDVI time-series plots to each polygon---evidence that helps distinguish maize from other crops by growth pattern. The user labels a small seed set: maize, other-crop, or ignore. Propagation expands the labels by finding similar polygons in embedding space. Dual modeling then scales: a lightweight classifier produces a probability surface over the ROI, masked to cropland. During quality control, uncertainty sampling prioritizes review in low-confidence regions. The session ends with an exportable map and labeled training polygons.
\end{userstory}

\begin{userstory}{Flood Damage Assessment}
\includegraphics[width=\linewidth]{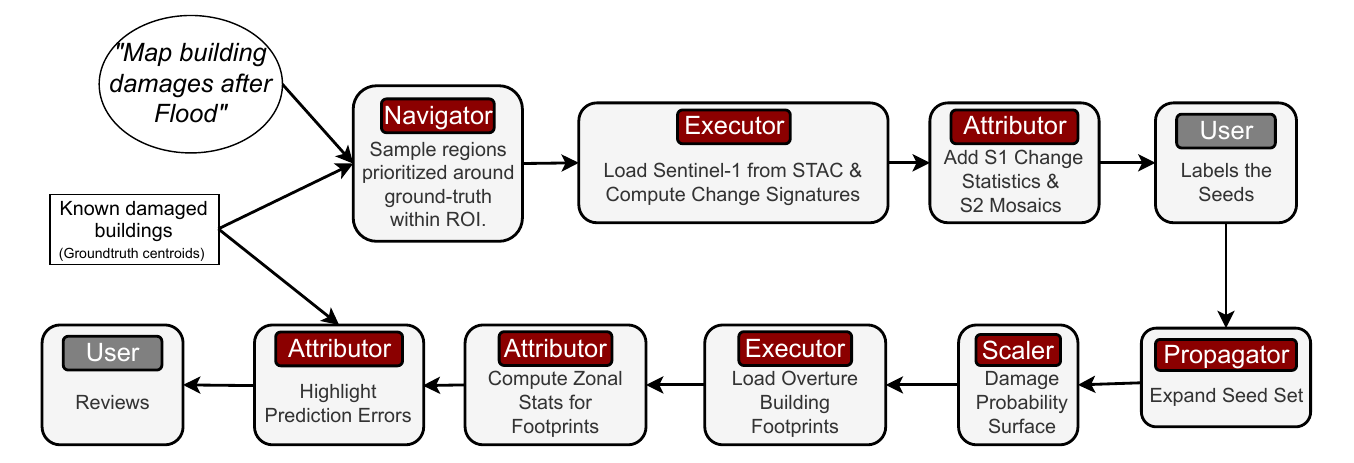}
\label{fig:flood_mapping}

A user needs a building-level damage map for Derna, Libya following the 2023 dam collapse---with a 24-hour deadline. They have centroid points for a few confirmed destroyed buildings but no systematic labels.

The user sets an ROI around the impacted corridor and a time window from flood onset to present. The agent loads Sentinel-1 SAR imagery for pre- and post-event periods, since optical imagery has heavy cloud cover. Working from SAR change signatures---strong backscatter changes, water extent patterns---the user creates seed labels for damaged vs.\ undamaged areas.

The agent enriches each seed with SAR change statistics and Sentinel-2 mosaic previews, making review easier. Ground-truth centroids anchor the process: the agent prioritizes sampling around them and checks consistency. Propagation expands the seed set. Dual modeling produces a damage probability surface.

Finally, the agent loads Overture building footprints and aggregates predictions per building. A quick validation checks what fraction of ground-truth destroyed-building centroids fall within predicted damaged buildings. The session ends with exportable artifacts: a building damage layer with scores, the underlying probability raster, and documentation of data sources and validation results.
\end{userstory}

\end{document}